\documentclass[conference]{IEEEtran}
\IEEEoverridecommandlockouts
\usepackage{cite}
\usepackage{amsmath,amssymb,amsfonts}
\usepackage{algorithmic}
\usepackage{graphicx}
\usepackage{textcomp}
\usepackage{xcolor}
\usepackage{hyperref}

\def\BibTeX{{\rm B\kern-.05em{\sc i\kern-.025em b}\kern-.08em
    T\kern-.1667em\lower.7ex\hbox{E}\kern-.125emX}}

\usepackage{multirow}
\usepackage{diagbox}
\usepackage{subcaption}
\DeclareMathOperator*{\argmax}{arg\,max}

\begin{document}

\title{Structure-Aware Hierarchical Graph Pooling using Information Bottleneck
}

\author{\IEEEauthorblockN{Kashob Kumar Roy, Amit Roy, A K M Mahbubur Rahman, M Ashraful Amin and Amin Ahsan Ali}
\IEEEauthorblockA{\textit{Artificial Intelligence and Cybernetics Lab}, \textit{Independent University Bangladesh}\\
\{kashobroy, amitroy7781\}@gmail.com, and \{akmmrahman, aminmdashraful, aminali\}@iub.edu.bd}}


\maketitle

\begin{abstract}
 Graph pooling is an essential ingredient of Graph Neural Networks (GNNs) in graph classification and regression tasks. For these tasks, different pooling strategies have been proposed to generate a graph-level representation by downsampling and summarizing nodes' features in a graph. However, most existing pooling methods are unable to capture distinguishable structural information effectively. Besides, they are prone to adversarial attacks. In this work, we propose a novel pooling method named as {HIBPool} where we leverage the Information Bottleneck (IB) principle that optimally balances the expressiveness and robustness of a model to learn representations of input data. Furthermore, we introduce a novel structure-aware Discriminative Pooling Readout ({DiP-Readout}) function to capture the informative local subgraph structures in the graph. Finally, our experimental results show that our model significantly outperforms other state-of-art methods on several graph classification benchmarks and more resilient to feature-perturbation attack than existing pooling methods\footnote{Source code at: \href{https://github.com/forkkr/HIBPool}{\color{blue}https://github.com/forkkr/HIBPool}}.
\end{abstract}


\section{Introduction}
Graph Neural Networks (GNNs) have seen tremendous success in learning meaningful representations from graph-structured data for various downstream tasks such as graph-classification~\cite{cangea2018towards,ranjan2020asap,lee2019self,xu2018powerful,gao2019graph,zhang2018end}, graph-based regression~\cite{sanyal2018mt,xie2018crystal}, and node classification~\cite{hamilton2017inductive,kipf2016semi}.

\begin{figure}[tb]
\centering
\includegraphics[width=0.99\columnwidth, height=0.6\columnwidth]{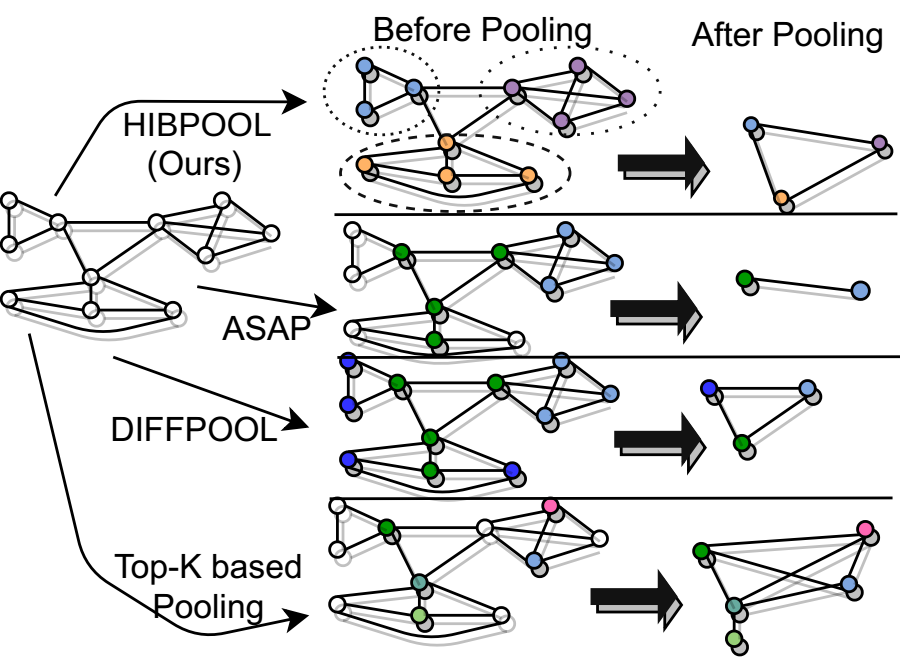}
\caption{Overview of pooling methods: HIBPool, DIFFPOOL (colors denotes local communities/clusters); ASAP (color denotes selected top k local clusters); Top-K based Pooling (colors indicates the set of selected top k nodes)}
\label{subg_conv_fig}
\end{figure}

In brief, graph classification aims to predict the label of an input graph by utilizing the given graph structure and node features. Inspired by CNNs, graph local pooling methods compute the graph-level representations through downsampling and summarizing the local structures and node features hierarchically. One line of research on pooling methods~\cite{gao2019graph,lee2019self,li2020graph} follows Top-$K$ node selection strategies. It involves setting different criteria to select a subset of top $K$ nodes with high scores based on the criteria to form a coarser graph for subsequent pooling layer (see Fig.~\ref{subg_conv_fig}). Another line of research~\cite{ying2018hierarchical,ranjan2020asap,bianchi2020spectral,bacciuk,luzhnica2019clique} focuses on capturing local substructures explicitly through partitioning the graph into several communities/clusters and finding one representative node for each community/cluster to form next layer coarser graph (see Fig.~\ref{subg_conv_fig}). 

However, recently, authors in~\cite{mesquita2020rethinking} ran experiments on several existing pooling methods raised questions about the efficacy of local pooling and concluded that the accuracy of graph classification methods based on existing local poolings are not superior to a simple global pooling method that computes a graph-level summary representation from all node features through a simple weighted summation or neural network directly. This paper investigates this claim further. There could be several factors responsible for the lack of performance improvement. Firstly, the existing methods might be incapable of learning important local community/subgraph structures. Secondly, the simple classification loss without regularization terms used in graph classification may promote learning homogeneous node representations across the whole graph
that naturally impose challenges for simple local pooling functions to learn informative local structures. Thirdly, irrelevant or redundant node features may also produce more homogeneous representations. We elaborate on these points below and describe the contributions of this paper.

First, the ability to define proper assignments of nodes to local community/clusters plays an essential role in learning important local structural information. Communities in a graph are groups of nodes with denser connections among them than connections with the rest of the graph, which may share common properties or play similar roles, or represent functional modules as a whole~\cite{fortunato2010community}. Community structures are ubiquitous and play essential roles in the topology and function of various graph networks such as biological networks~\cite{sah2014exploring}, social networks~\cite{girvan2002community} etc. For example, the class label of a protein is enzyme or non-enzyme, which depends on not only features of amino-acids but also the community structures in amino-acids interaction networks~\cite{gaci2011community} or motifs of amino-acids in its amino-acid sequence~\cite{yao2003accurate}. Besides, these structures are important to identify the functional modules as well as to capture the hierarchical feature information in graphs.  Different local pooling algorithms use different local substructures to pool from and thus differ in their ability to capture local information and produce robust graph representations. In recent years, few papers adapted different partitioning approaches, which are very popular in complex network analysis.~\cite{luzhnica2019clique} proposes pooling operation using clique i.e., each of its nodes is connected to
all the others.~\cite{bacciuk} introduces K-plex cover into graph pooling in which each
node is adjacent to all other nodes within a subgraph except at most k of them. Authors in~\cite{bianchi2020spectral} leverage the formulation of the Min-Cut problem as a regularization term with classification loss to compute cluster assignment of nodes. However, these algorithms have several shortcomings. The methods take the number of communities or their sizes as a parameter and only consider internal cohesion within subgraphs~\cite{fortunato2010community,newman2006modularity}. It is quite expected that Top-K selection based pooling methods~\cite{gao2019graph,lee2019self,li2020graph} might ignore the topological structure of the graph. In Fig.~\ref{subg_conv_fig} we can see that some nodes (w/o color) may have low scores, but they have an important role in topological structures. More structure-aware methods such as DIFFPOOL~\cite{ying2018hierarchical} hierarchically learns the cluster assignment matrix. ASAP~\cite{ranjan2020asap}, on the other hand, defines all possible local clusters consisting of a target node and its 1-hop neighborhood and selects top $k$ local clusters depending on the feature-based fitness scores. Further, we analyze the consistency of local clusters learned by DIFFPOOL and ASAP with respect to the set of optimal communities discovered by Louvain~\footnote{\label{louvain}Note that we never know the true community structures.  Hence  we  choose  the  most  popular  and  widely accepted algorithm for detecting the communities~\cite{blondel2008fast}}\cite{blondel2008fast} from graph as baselines. 
\begin{table}[h]
\small
\centering
\begin{tabular}{c|cc|cc|cc}
\hline
\multicolumn{1}{c}{\textbf{Datasets}} & \multicolumn{2}{c}{\textbf{ENZYMES}} & \multicolumn{2}{c}{\textbf{D\&D}} & \multicolumn{2}{c}{\textbf{PROTEINS}} \\ \hline
 Model & AMI & ARI & AMI & ARI & AMI & ARI \\ \hline
DIFFPOOL       & 0.24  & 0.18 & 0.04  & 0.02  & 0.10  & 0.04    \\
ASAP       & 0.16  & 0.10 & 0.12  & 0.06  & 0.14  & 0.08    \\
\hline
\end{tabular}
\caption{Consistency of learned communities w.r.t the communities generated by Louvain$^{\ref{louvain}}$ in original graphs: AMI - Adjusted Mutual Information and ARI - Adjusted Rand Index. The higher value indicates better consistency.}
\label{tab:community-comparison}
\end{table}

In Table~\ref{tab:community-comparison}, we can observe that the values of Adjusted Mutual Information (AMI) and Adjusted Rand Index (ARI) are very low, which demonstrates that they fail to capture community structures in graphs. However, most state-of-the-art community detection methods find communities based on the maximization of modularity~\cite{blondel2008fast,newman2006modularity,rahiminejad2019topological} and aim to identify the community structures of the graph as a whole. We observed that the effectiveness of modularity-based community detection algorithms to capture topological and functional substructures in graph classification tasks had not been investigated well.

\begin{figure}[!tb]
\centering
\includegraphics[width=0.9\columnwidth]{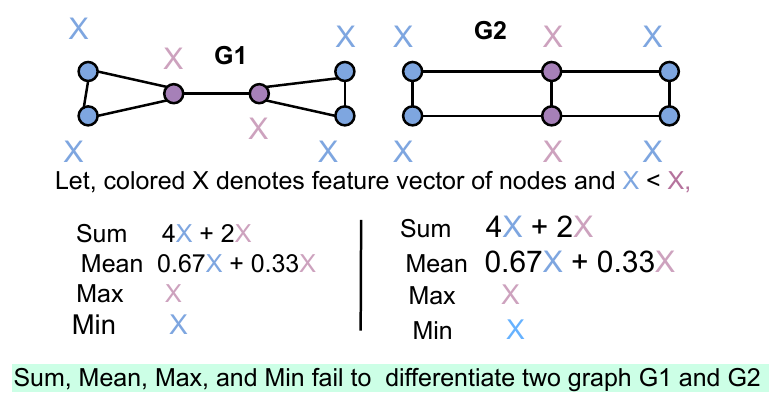}
\caption{With identical feature vector for all same colored nodes simple readout like Sum, Mean, Max, Min etc. can not distinguish the substructures G1 and G2. To discriminate G1 and G2, centrality measures e.g. clustering coefficient can be incorporated to design sophisticated aggregation function that we have discussed in our proposed section.}
\label{ag_moti_fig}
\end{figure}

Second, simple pooling functions such as Sum, Mean, Max, Min, etc., fail to obtain distinguishable representations of different substructures with homogeneous node feature vectors (as shown in Fig.~\ref{ag_moti_fig}). Authors in ~\cite{mesquita2020rethinking} state that multiple convolutions on node features encourage to learn over-smoothed representation across their neighborhoods.~\cite{xu2018powerful,corso2020principal} discussed the limited expressive power of simple aggregation functions to differentiate messages from the neighborhood in the node classification context. Therefore, we conduct rigorous experiments to investigate the limitations of simple pooling functions and design a novel discriminative pooling readout ({DiP-Readout}) function that addresses the limitations.
Moreover, as the topological features characterize the structures and dynamics of networks and are necessary to identify different categories of structures~\cite{costa2007characterization}, we incorporate the topological features of nodes into the {DiP-Readout} function to distinguish the different structures of local communities.

Third, Information Bottleneck (IB) suggests that an optimal learning method should maximize the mutual information between latent representation and the target label to make prediction accurate while minimizing the mutual information between the latent representation and the input to promote generalization ability~\cite{tishby2000information,tishby2015deep}. GIB~\cite{wu2020graph} introduces IB to regularize the structural and feature information in the node classification task. Similarly, for graph classification, IB  may promote the pooling function to learn the informative pooled representation of local communities with minimal redundancy, which would lead to distinct representations. Therefore, in this work, we leverage IB for graph pooling, which encourages the pooled representations for each community to be maximally informative about the target graph label while discouraging to accumulate irrelevant information. Hence, it helps to achieve minimal sufficient graph-level summary representation.

In summary, our key contributions are as follows -
\begin{itemize}
    \item We propose a novel graph pooling (HIBPool) method to capture structural and feature information from communities/local substructures in a hierarchical manner to achieve more comprehensive graph-level representations.
    \item We introduce more expressive structure-aware readout function {DiP-Readout} to differentiate discriminative substructures/communities. 
    \item We leverage the IB principle into pooling method and demonstrate its effectiveness in improving the pooling capability on graph classification. 
    \item We empirically show that our model HIBPool significantly outperforms the state-of-art approaches on six graph classification benchmarks.
     \item We empirically show that our model HIBPool  is resilient to the feature adversarial attacks.
\end{itemize}

\section{Background and Related works}

\subsection{Graph Neural Networks (GNNs).} In recent years, a plethora of research has been conducted to learn either node-level or graph-level meaningful representations through GNNs~\cite{zhou2020graph}. Several approaches~\cite{kipf2016semi,hamilton2017inductive,velivckovic2017graph,xu2018powerful} follow a neighborhood aggregation scheme, where the representations of a node is computed through iterative aggregation and transformation of neighbor-node feature-information. They are also called as Message-Passing Networks (MPN) since nodes aggregate messages from neighbors through edges. Those are inherently flat and ineffective in learning hierarchical structural information in a graph.  

\subsection{Non-topological Graph Pooling: } The majority of graph pooling methods perform pooling to reduce the graph structure based on node embeddings that hardly reflect graph topological information. To learn hierarchical representations of graphs, DIFFPOOL~\cite{ying2018hierarchical} aims to learn cluster assignment matrix from node features and map the nodes to different clusters and thus generate coarser graph for the next layers. By utilizing graph convolution SAGpool~\cite{lee2019self} determines self-attention scores, and the top $k$ portion of the nodes are retained in the next pooling layer.
Similarly, inspired by the encoder-decoder based U-Net architectures widely used for image augmentation in CNNs, Graph U-Nets~\cite{gao2019graph} proposed pooling operation gPool to adaptively select nodes based on their scalar projection values on a trainable projection vector and inverse operation gUnpool for upsampling. ASAP~\cite{ranjan2020asap} proposed self-attention to determine the importance of each node and learn the soft cluster assignment while GXN~\cite{li2020graph} introduce VI-Pool to select the informative set of vertices that maximizes the mutual information between node and neighborhood features. All of those approaches are local pooling approach that pool graphs in a hierarchical way. In contrast, global pooling methods use summation or neural networks to pool all node embeddings into a global summary representation.

\subsection{Topological Graph Pooling}
Recent few works exploit graph topology to pool the graph.~\cite{luzhnica2019clique} pool the graph using cliques, i.e., every node is connected with all other nodes in a group. The hard constraint of clique potentially destroys the relationships among nodes in a graph. Consequently, there are few relaxed concepts such as $n$-cliques, $n$-clans, $k$-plexes, etc., to partition the graph.~\cite{bacciuk} proposes a pooling method built on the concept of graph $k$-Plexes, i.e., pseudo-cliques where every node has direct ties to $n-k$ members in a group of size $n$. In other words, a clique is equivalent to $1$-plex. Still, the constraint of $k$-plex is too strict to find all interesting structural relationships. It is naturally impossible to determine a fixed value of $k$ to find out all structures.~\cite{bianchi2020spectral} utilize the formulation of the $k$-way Min-Cut problem as a regularization term to guide the pooling function towards capturing the local structures in the graph. This partitioning algorithm depends on the value of hyper-parameter $k$. These limitations render the above methods unsuitable for capturing community structures in a graph.       
\subsubsection*{Community-based Graph Pooling}
Community detection is of great importance in various disciplines such as sociology, biology, computer science, etc., where systems are often represented as graph networks~\cite{fortunato2010community, newman2006modularity}. Communities uncover important topological and functional features in graphs~\cite{rahiminejad2019topological}. These topological and functional feature information can enrich the embeddings of graphs. Newman-Girvan modularity in Equation~\ref{eq:modularity}, the most popular global quality function, is introduced to detect the community structures of the graph~\cite{newman2004finding,newman2006modularity} where the high values of modularity indicate the good communities. The methods based on the modularity maximization~\cite{newman2004fast,blondel2008fast} aims to identify
groups of nodes that are more densely connected than one would expect from a statistical
null model of the network. These methods do not require the number of communities or the size of communities as predefined hyper-parameter and can balance the groups' size in terms of their total connectivity. Louvain algorithm~\cite{blondel2008fast} is likely the best method to find communities, especially in biological graphs~\cite{rahiminejad2019topological}, and faster. However, the existing methods fail to explicitly leverage the advantage of inherent community structure into graph pooling that is purely topological, nonparametric, and more interpretable than non-topological pooling methods as there are no learnable parameters. Therefore, in this paper, we introduce community-based hierarchical pooling by using Louvain algorithm~\cite{blondel2008fast} to detect the optimal set of communities.

\subsection{Preliminaries:}
We consider a graph $\mathcal{G} = (V, A, X)$ with $N= |V|$ where $V$ and $A$ are the set of nodes and adjacent matrix respectively. $X \in R^{N \times f}$ presents the node feature matrix. Graphs are associated with a class label  $y$ from a defined set of labels, $Y=\{1,\dots,c\}$. The goal of graph classification task is to learn a mapping function $F(\mathcal{G}): \mathcal{G} \longrightarrow Y$ and predicts the labels of unseen graphs.

\begin{figure*}[tb]
\centering
\includegraphics[width=0.98\textwidth]{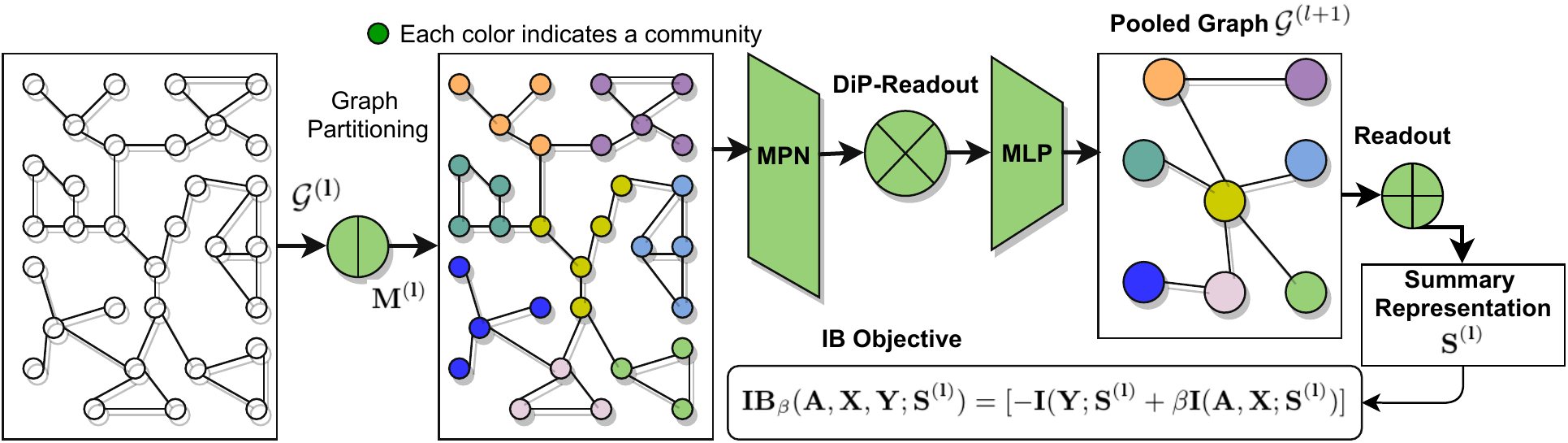}
\caption{To determine different substructures present in graph data, the input graph is partitioned into communities where  intra-community nodes are represented with same color and they are densely connected than inter-community nodes. After aggregating features from neighborhood using Messaging Passing Network (MPN) we apply structure-aware DiP-Readout to obtain distinguishable representations of communities with different substructures while the pooled graph of next layer is created with the edges connecting different communities. Summary representation is obtained by a readout function after the final layer. Optimizing with IB principal ensure minimal redundancy from graph data with sufficient information to predict the class labels.}
\label{pooling_layer_fig}
\end{figure*}

\section{HIBPool: Hierarchical IB-guided Pooling}
We apply our proposed HIBPool on each layer graph $\mathcal{G}^{(l)} = (V^{(l)}, A^{(l)}, X^{(l)})$ and form coarser pooled graph $\mathcal{G}^{(l+1)} = (V^{(l+1)}, A^{(l+1)})$ for next layer $(l+1)$. Here, $V^{(l)}$ and $A^{(l)}$ are the set of nodes and adjacent matrix respectively, $N^{(l)}= |V^{(l)}|$ at layer $l$. $\mathcal{G}^{(1)}$ denotes the given input graph. Figure~\ref{pooling_layer_fig} shows the brief overview of HIBPool that is described elaborately in following sections. 

\subsection{Community Detection}
Community structures unveil interesting relationships among nodes in a graph. Intra-community nodes are densely interconnected and probably share common properties and/or play similar roles within the graph. Analogous to standard CNNs, we consider local communities in the graph as local patch and compute pooled latent representation for each community (patch). We employ Louvain algorithm~\cite{blondel2008fast} to partition the graph into communities by optimizing the modularity $\mathcal{Q}$ (in Eq.~\ref{eq:modularity})  in a bottom up fashion where $d^{(l)}_{v}$ is degree of node $v$, $| A^{(l)} |$ is total number of edges, $\delta_{uv} = 1$ when both nodes $u$ and $v$ are in the same community otherwise $0$. 
\begin{equation}
    \mathcal{Q}(P^{(l)}) = \frac{1}{2|A^{(l)}|} \sum_{u, v} [A^{(l)}_{uv}-\frac{d^{(l)}_{u}d^{(l)}_v}{2|A^{(l)}|}]\delta_{uv}
    \label{eq:modularity}
\end{equation}
 We finds a set of communities, $P^{(l)} = \{P^{(l)}_1, P^{(l)}_2, \dots \}$ in graph $\mathcal{G}^{(l)}$ under the optimal partitioning that would maximize the modularity as following,
\begin{equation}
    P^{(l)} = \argmax_{\mathcal{P}_{i} \in \mathcal{P}}\mathcal{Q}(\mathcal{P}_{i})
\end{equation}
where the set of all possible partitioning is $\mathcal{P} =\{ \mathcal{P}_1, \mathcal{P}_2, \dots \} $.
Afterwards, community mapping matrix $M^{(l)} \in R^{N^{(l+1)}\times N^{(l)}}$ is computed by following the Eq.~\ref{eq:assigment} where $P^{(l)} =\{P^{(l)}_1, P^{(l)}_2, \dots\}$ and $N^{(l+1)}=|P^{(l)}|$. In Figure~\ref{pooling_layer_fig}, different colors indicate different community structures.
\begin{equation}
    M^{(l)} = \{M^{(l)}_{(i, v)} = 1| \forall v \in P^{(l)}_i, P^{(l)}_i \in P^{(l)}, v \in V^{(l)}\}
    \label{eq:assigment}
\end{equation}
We use the mapping $M^{(l)}$ to compute the latent representations for each community and to construct coarser graph for next pooling layer. In Figure~\ref{pooling_layer_fig}, each colored node in the pooled graph $\mathcal{G}^{(l+1)}$ corresponds to a particular community in $\mathcal{G}^{(l)}$.
\subsection{Feature Propagation}
Before applying DiP-Readout to compute the representations of each community, we apply a message passing network (MPN) to propagate features across neighbor nodes as follows:
\begin{equation}
    Z^{(l)} = ReLU(X^{(l)}*W^{(l)}_{1}+A^{(l)}X^{(l)}W^{(l)}_{2}) \in R^{N^{(l)}\times h}
    \label{eq:mpn}
\end{equation}
where $W_1^{(l)}$, $W_2^{(l)}$ are weights at layer $l$ and $h$ -  output dimension. 

\begin{figure*}[tb]
\centering
\includegraphics[width=0.98\textwidth, height=0.3\textwidth]{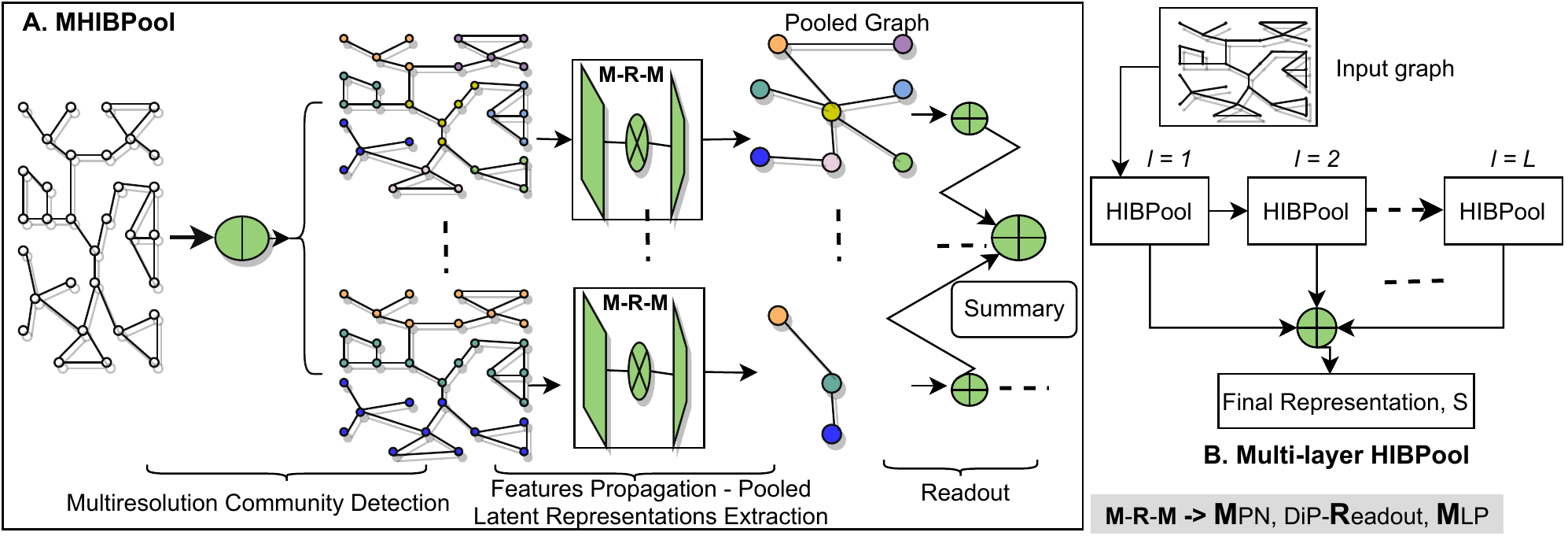}
\caption{Overview of A) MHIBPool with varying the value of resolution parameter $\alpha$; B) Multi-layer HIBPool}
\label{model_fig}
\end{figure*}

\subsection{Discriminative Pooling Readout, (DiP-Readout)} In Figure~\ref{subg_conv_fig}, we show that simple readout functions fail to distinguish different structures. Thus, the simple readout functions limit the generalization ability of a model for graph classification.  For the node classification tasks, Corso et al.~\cite{corso2020principal} demonstrates that a hybrid aggregator PNA consisting of four simple aggregators with three degree-scalers can effectively increase the ability to capture different neighborhood structures. Nevertheless, it is to be noted that directly incorporating PNA as a readout function is not enough to discriminate different substructures. We can see that the degree-based scaling of PNA could not differentiate two substructures in Figure~\ref{ag_moti_fig}. Therefore, we utilize the topological features such as degree, clustering coefficient, and betweenness centralities, etc., to design a structure-aware pooling readout that generates pooled representation for each community while preserving different substructures. To define the discriminative pooling readout {(DiP-Readout)}, let $C \in R^{|V^{(l)}|\times \kappa}$ where $\kappa$ is the number of centralities that we consider here. Then we normalize the centralities as follows,
\begin{equation}
    \forall v \in V^{(l)}, i = \delta_v, \widehat{C}[v,j] = \frac{exp(C[v,j])}{\sum_{u \in P^{(l)}_{i}}exp(C[u,j])}
\end{equation}
where, $\delta_v$ denotes the community membership of node v.
Next, we compute scaled-embeddings of nodes by multiplying centralities with the embeddings $Z^{(l)}$. We get different scaled-embeddings of nodes for each centrality. We concatenate $\kappa$ scaled-embeddings with no scaled-embeddings in following,
\begin{equation}
    \widehat{Z}^{(l)} = Z^{(l)} || (\widehat{C} \odot Z^{(l)}) \in R^{|V^{(l)}|\times h(\kappa+1)}
\end{equation}
where $\odot$ - broadcasted hadamard product, $||$ - concatenation, $|\cdot|$ - cardinality. 

It is to be noticed that each community is a multiset of node feature vectors. In order to discriminate between multisets of features, we compute a statistic over node features within a community such as Sum, Mean, Max, Min as follows, 
\begin{equation}
\begin{split}
    &Z_{sum} =  M^{(l)}\widehat{Z}^{(l)} \in R^{|P^{(l)}|\times h(\kappa+1)} \\
    &Z_{mean} = D^{-1}M^{(l)}\widehat{Z}^{(l)} \in R^{|P^{(l)}|\times h(\kappa+1)} \\
    &Z_{max}[i,:] = \max_{j \in p_{i}}(\widehat{Z}^{(l)}_{j}) \in R^{|P^{(l)}|\times h(\kappa+1)} \\
    &Z_{min}[i,:] = \min_{j \in p_{i}}(\widehat{Z}^{(l)}_{j}) \in R^{|P^{(l)}|\times h(\kappa+1)}
\end{split}
\end{equation}
where $D$ is diagonal degree matrix of $M^{(l)}$, $h$ -  output dimension, $||$ - concatenation, $|\cdot|$ - cardinality.

After computing the statistics for communities, we concatenate and pass them through MLP to compute discriminative representations of each community as, 
\begin{equation}
    H^{(l)} = MLP(Z_{sum}||Z_{mean}||Z_{max}||Z_{min})\in R^{|P^{(l)}|\times 2h}
\end{equation}

Afterwards, we compute the summary representation of the graph at pooling layer $l$ as follows,
\begin{equation}
    S^{(l)} = \frac{1}{|P^{(l)}|} \sum_{i=1}^{|P^{(l)}|} H^{(l)}[i, :h] \in R^{1\times h}
\end{equation}
\subsection{Pooled Graph}
We compute $\mathcal{G}^{(l+1)} = (A^{(l+1)}, X^{(l+1)} )$ and pass it as an input for next pooling layer as following,
\begin{equation}
\begin{split}
    A^{(l+1)} &= M^{(l)}\widehat{A}^{(l-1)}M^{(l)T} \in R^{|P^{(l)}|\times|P^{(l)}|} \\
    X^{(l+1)} &= H^{(l)}[,:h]
    \label{eq:new_adjacency}
\end{split}
\end{equation}
where $\widehat{A}^{(l)} = A^{(l)} + I$, $^T$ denotes transpose operation. Equation~\ref{eq:new_adjacency} ensures the connectivity of pooled graph where any two pooled node i and j are connected if there is any edge between node pair $u \in P^{(l)}_i$ and $v \in P^{(l)}_j$.
\subsection{Deriving Information Bottleneck for Pooling}
To incorporate IB principle~\cite{wu2020graph} into pooling, it requires pooled representations $S^{(l)}$ to minimize the redundant information from ($A^{(l)}, X^{(l)}$) and maximize the information to predict class label $Y$. The intractability of general IB formulation leads us to approximate our objective in Equation~\ref{eq:ib_objective} with some additional assumptions. Here, we rely on the community-dependence assumption: each pooled representation $H^{(l)}_{i}$ only depends on respective community structure, $P^{(l)}_i$ and intra-community node features, $\{ X_{v}^{(l)} | v \in P^{(l)}_{i} \}$ while being independent of the rest of the graph.
\begin{equation}
    IB_{\beta}(A^{(l)},X^{(l)},Y; S^{(l)}) = [ - I(Y;S^{(l)}) + \beta I(A^{(l)},X^{(l)};S^{(l)})]
    \label{eq:ib_objective}
\end{equation}
With our community-depedence assumption, we can simply estimate $I(A^{(l)},X^{(l)};S^{(l)})$ as follows,
\begin{equation}
\small
    I(A^{(l)},X^{(l)};S^{(l)}) \longrightarrow \sum^{|P^{(l)}|}_{k=1} I(A^{(l)},X^{(l)};H^{(l)}_{k})
\end{equation}

To instantiate the objective, following GIB~\cite{wu2020graph}, we set an upper bound of  $I(A^{(l)},X^{(l)};H^{(l)}_{k})$, and estimate the upper bound by using the sampled $H^{(i)}_{X,j} \sim \mathcal{N}(\mu_{i,j}, \sigma^{2}_{i,j})$ where $\mu_{i, j}=H^{(i)}_{j}[j,:h]$ and $\sigma^{2}_{i,j} = H^{(i)}_{j}[j, h:]$ as,
\begin{equation}
\begin{split}
    &I(A^{(l)},X^{(l)};H^{(l)}_{k}) \rightarrow \log\frac{P(H^{(l)}_{X,k}|A^{(l)},X^{(l)})}{Q(H^{(l)}_{X,k})} \\
    &= \log \Phi (H^{(l)}_{X,k}; \mu_{l, j}, \sigma_{l, j}^{2})\\
    &- \log(\sum_{v \in P_{i}} \Phi ( H^{(l)}_{X,k}; \mu_{l-1, v}, \sigma^{2}_{l-1, v})) \\
    \label{eq:xib}
\end{split}
\end{equation}

To estimate the upper bound of $I(A^{(l)},X^{(l)};H^{(l)}_{k})$, we assume for any node $v$, the given input feature $X_v \sim \mathcal{N}(\mu_{0,v}=0,\sigma^{2}_{0,v}=1)$. 

Again, to estimate $I(Y;S^{(l)})$, we use the cross-entropy loss i.e.,
\begin{equation}
\small 
    I(Y;H^{(l)}) \longrightarrow - \sum_{i=1}^{|Y|} Y_{i} \log(S^{(l)}W_{out})
    \label{eq:XEntropy}
\end{equation}
Finally, plugging the Equation~\ref{eq:xib} and Equation~\ref{eq:XEntropy} into Equation~\ref{eq:ib_objective}, we utilize the benefit of IB principle to learn minimally redundant maximally informative graph-level representations through training model by optimizing the objective function.

\subsection{Multi-layer HIBPool}
Similar to Figure~\ref{model_fig}(B), we stack $L$ HIBPool layers to compute the graph-level representation hierarchically. For each pooling layer $l$, we execute HIBPool to compute the summary representation $S^{(l)}$. Afterward, we concatenate the summary representations of all layer and take the weighted summation as,
\begin{equation}
    S = \sum_{i=1}^{L} W_fS^{(l)}
\end{equation} 
$S$ is considered as the final representation and is used to predict the label of input graph.

\subsection{MHIBPool: Multiresolution HIB Pooling}
We can see in Equation~\ref{eq:modularity} that only node pairs belonging to the same community contribute to the sum, so we can group these contributions of intra-community nodes together and rewrite the sum over the node pairs as a sum over the community, as follows,
\begin{equation}
    Q(P^{(l)}) = \frac{1}{2}\sum^{N^{(l)}}_{c=1}[\frac{e^{(l)}_c}{|A^{(l)}|} - (\frac{(d^{(l)}_c)^2}{2(|A^{(l)}|)^2})]
    \label{eq:com_modularity}
\end{equation}
where $N^{(l)}$ is the total number of communities, $e^{(l)}_c$ is the total number of intra-community edges, and $d^{(l)}_e$ is the sum of the degree of intra-community nodes. 

To address the Louvain algorithm's resolution limitation, we use multi-scale modularity~\cite{pons2011post} to detect the multiresolution communities. The multi-scale modularity can be defined as,
\begin{equation}
    Q(P^{(l)}) = \sum^{N^{(l)}}_{c=1}[\alpha \frac{e^{(l)}_c}{|A^{(l)}|} - (1-\alpha)(\frac{d^{(l)}_c}{2(|A^{(l)}|)})^2]
    \label{eq:multi_modularity}
\end{equation}
In Equation~\ref{eq:multi_modularity}, $0< \alpha <1$ is the resolution parameter. We see that for $\alpha = \frac{1}{2}$, the equation is equivalent to~\ref{eq:com_modularity}. Therefore, we can find multiresolution communities by varying the value of $\alpha$ (see Figure~\ref{model_fig}(A)). Similar to multi-layer HIBPool in Figure~\ref{model_fig}(B), the multi-layer MHIBPool consists of multple consecutive layers of MHIBPool. 


\section{Experimental Analysis}
In this section, we validate the efficacy of our model by conducting extensive experimental analysis. We present the datasets and experimental setup, comparison with baselines, the effectiveness of the DiP-Readout function, and the effect of adding the IB principle. Finally, we show the resilience of our model against feature perturbations.

\subsection{Datasets and Experimental Setup} 
To investigate our research findings mentioned in the introduction section and evaluate our proposed model's efficacy, we have performed experiments on six popular graph classification benchmark datasets: ENZYMES, DD, PROTEINS, NCI1, NCI109, and FRANKENSTEIN~\cite{ranjan2020asap}. Those datasets are bioinformatics datasets and used for performance comparison in several state-of-the-art pooling methods e.g. DIFFPOOL~\cite{ying2018hierarchical}, SAGPool~\cite{lee2019self}, Graph U-Net~\cite{gao2019graph}, and GXN~\cite{li2020graph} etc. In all the experiments, we followed 80\%, 10\%, and 10\% split for training, validation, and test set, and then performed 10-fold cross-validation, and reported the average accuracy for evaluation. 
We perform fine-tuning for the hyperparameter of IB term $\beta = 0.01$. For all the experiments, we use Adam optimizer and set the learning rate as 0.01. We set $L=2$ for all experiments. For our baseline models, we follow the exact experiment setup that is mentioned in their manuscript.


\begin{table*}[]
\small
\begin{tabular}{c|c|c|c|c|c|c}
\hline
\backslashbox{\textbf{Model}}{\textbf{Dataset}}                                               & \textbf{ENZYMES} & \textbf{D\&D} & \textbf{PROTEINS} & \textbf{NCI1} & \textbf{NCI109} & \textbf{FRANKENSTEIN}                                  \\ \hline
\textbf{DIFFPOOL~\cite{ying2018hierarchical}}                                                                    & \underline{62.53 ± 2.74}            & 66.95 ± 2.41  & 68.20 ± 2.02      & 62.32 ± 1.90  & 61.98 ± 1.98    & 60.60 ± 1.62                                           \\

\textbf{KPLEXPOOL~\cite{bacciuk}}                             & 39.67 ± 7.52  & 77.76 ± 2.92      & 75.11 ± 2.80  & \underline{79.17 ± 1.73}    & -        & -                                               \\

\textbf{SAGPool~\cite{lee2019self}}                                                                     & 43.99 ± 4.23     & 76.45 ± 0.97  & 71.86 ± 0.97      & 67.45 ± 1.11  & 67.86 ± 1.41    & 61.73 ± 0.76                                           \\

\textbf{Graph U-Net~\cite{gao2019graph}}                                                                 & 50.21 ± 5.53     & 81.34±3.23    & 76.78 ± 4.23      & -             & -               & -                                                      \\ 
\textbf{MinCutPool~\cite{bianchi2020spectral}}                                                                 & -     & 80.8±2.3    & 76.5±2.6      & -             & -               & -                                                      \\ 
\textbf{ASAP~\cite{ranjan2020asap}}                                                                        & 29.43  ± 2.34                & 76.87 ± 0.7   & 74.19 ± 0.79      & 71.48 ± 0.42  & \underline{70.07 ± 0.55}    & \underline{66.26 ± 0.47}                                           \\ 
\textbf{GXN~\cite{li2020graph}}                                                                        & 57.50  ± 6.1                & \underline{82.68 ± 4.1}   & \underline{79.91 ± 4.1}      & -  & -    & -                                           \\ \hline \hline

\textbf{\begin{tabular}[c]{@{}c@{}}HPool (Mean) w/o IB\end{tabular}} & 60.0 ± 2.3       & 81.32 ± 2.12  & 77.12 ± 0.34       & 74.32 ± 0.85  & 76.02 ± 1.94     & 62.09 ± 1.51                                           \\ 

\textbf{\begin{tabular}[c]{@{}c@{}}HPool (DiP-Readout) w/o IB\end{tabular}}  & 61.33 ± 1.66      & 79.23 ± 1.43  & 78.03 ± 2.14       & 77.37 ± 0.73  & 75.05 ± 1.69    & 64.66 ± 2.39                                           \\ 

\textbf{\begin{tabular}[c]{@{}c@{}}GlobalPool w/ IB\end{tabular}}    & 63.3 ± 1.95            & 80.66 ± 0.98         & 78.45 ±1.5             & 76.18±1.34          & 75.1 ± 1.24            & 63.78 ± 1.21                                                 \\ 

\textbf{\begin{tabular}[c]{@{}c@{}}RandomPool w/ IB\end{tabular}}    & 62.0 ± 2.34             & 77.27 ± 3.3        & 76.28 ± 2.21            & 74.20 ± 1.95       & 72.39 ± 2.5          & 66.12 ± 2.43                                                \\ 

\textbf{\begin{tabular}[c]{@{}c@{}}HIBPool (Mean)\end{tabular}}     & 71.67 ± 1.12        & 82.95 ± 1.56          & 81.98  ± 0.48            & 79.80 ± 0.69  & 77.73  ± 1.73     & \begin{tabular}[c]{@{}c@{}}71.42  ± 1.43\end{tabular} \\ 

\textbf{\begin{tabular}[c]{@{}c@{}}HIBPool\end{tabular}}     & \textbf{73.33 ± 1.8}       & \textbf{83.05  ± 0.34} & \textbf{82.88  ± 0.23} & \textbf{83.45 ± 0.73}  & \textbf{79.66 ± 0.72}     & \begin{tabular}[c]{@{}c@{}}\textbf{73.50 ±  0.23}\end{tabular} \\ \hline

\textbf{\begin{tabular}[c]{@{}c@{}}MHIBPool\end{tabular}}  & 75.0 ± 1.20           &    83.20 ± 1.27           & 80.35 ± 0.31  & 83.48 ± 0.90          & 79.93 ± 1.43           & 71.42 ± 0.53                    \\ \hline
\end{tabular}
\caption{Comparison with Baselines: '-' denotes that results are not publicly available.}
\label{tab:baseline_comparion}
\end{table*}

\subsection{Other Variants of HIBPool}
\begin{itemize}
    \item HPool (Mean) w/o IB - It finds communities and apply Mean readout to compute one higher-order representation for each community through optimizing simple classification loss.
    \item HPool (DiP-Readout) w/o IB - It employs \textit{DiP-Readout} instead of Mean readout function.
    \item GlobalPool w/ IB - It applies one-layer MPN and then takes the mean of all node embeddings to compute the graph-level representation. It utilizes the IB principle to learn parameters.
    \item RandomPool w/ IB - It replaces the community assignment of HIBPool by random matrix $M \sim \mathcal{N}(0, 1)$.

    \item HIBPool (Mean) - It replaces \textit{Dip-Readout} function of HIBPool by Mean pooling function.
\end{itemize}
\subsection{HIBPool Consistently Outperforms Baselines}
We compare HIBPool and its variants to recent state-of-arts methods including DIFFPOOL~\cite{ying2018hierarchical}, KPLEXPOOL~\cite{bacciuk}, SAGPool~\cite{lee2019self}, Graph U-Net~\cite{gao2019graph}, MinCutPool~\cite{bianchi2020spectral}, ASAP~\cite{ranjan2020asap} and GXN~\cite{li2020graph}. 
Table~\ref{tab:baseline_comparion} demonstrates the superior performance of our model HIBPool on all benchmark datasets for graph classification. We can see that simple HPool performs better than all baselines on 3 out of 6 datasets and very close to GXN (on D\&D, PROTEINS) and ASAP(on FRANKENSTEIN). It indicates that still community/local subgraph structural information promotes learning more informative graph-level representation. In contrast, learned assignment matrix~\cite{ying2018hierarchical} and Top-K selection strategies could not capture structural information explicitly. Again, the impressive performance of GlobalPool with IB over state-of-art methods on some benchmarks indicates that IB leads to learning more informative graph-level representation with minimal redundancy. Even HIBPool (Mean), which has the limited potential to capture local structural information, still outperforms all existing methods, HPool, and GlobalPool by a reasonable margin. This clearly indicates that both community structural information and IB regularization lead to attaining better accuracy. Furthermore, the structure-aware DiP-Readout function in HIBPool has more expressive potential to learn distinguishable representation for communities, even with homogeneous node features. Therefore we can conclude that several key factors, such as i) preserving structural information explicitly, ii) capturing discriminative community structures, and iii) minimizing the redundancy, promote HIBPool to learn sufficient informative representations, which leads to outstanding performance on all benchmarks. 

\subsection{Simple Pooling function is Not Enough}
State-of-art methods apply multiple convolutions to learn node embedding by utilizing graph structures. \cite{mesquita2020rethinking} showed that applying multiple convolutions may result in homogeneous node embeddings. It limits the effectiveness of simple pooling functions such as Sum, Mean, Max, Min, etc., to learn meaningful local community/subgraph structures (as shown in Fig.~\ref{ag_moti_fig}).
In Table~\ref{tab:baseline_comparion} HIBPool consistently shows greater average accuracy than HIBPool (Mean) on all six datasets. It implies that DiP-Readout is more expressive to learn meaningful local community structures than simple Mean function. However, HPool (Mean) shows competitive performance to HPool (DiP-Readout), which indicates that under simple classification loss, the DiP-Readout function suffers from redundant information. It may overshadow the expressive power of DiP-Readout. On the other hand, IB in Eq.~\ref{eq:ib_objective} promotes DiP-Readout to minimize redundant information by imposing constraints on the preserved information of input data while maximizing the preserved relevant information to predict the target.

\subsection{IB Principle Strengthens Local Pooling.}
To demonstrate the efficacy of Local Pooling to learn graph-level representation, we compare HIBPool to two other variants GlobalPool where the global mean pooling is performed, and RandomPool, where the random cluster assignment has been used ignoring intrinsic community structures in the graph. All three approaches are optimized with the supervised loss along with the IB principle. \cite{mesquita2020rethinking} concludes that graph convolutions is the main reason behind the success of local pooling approaches and shows that pooling does not play a significant role in performance improvement over global pooling. Table~\ref{tab:baseline_comparion} demonstrates that HIBPool outperforms both GlobalPool and RandomPool. These results indicate that the ability to capture the local/community structural information through local pooling is essential to produce better informative and distinguishable representations of graphs for classification. Here, IB promotes our local pooling method to capture this important structural information, resulting in superior performance.     

\begin{table}[]
\centering
\scalebox{0.9}{
\begin{tabular}{c|c|c|c|c}
\hline
\textbf{Models}
& \textbf{\begin{tabular}[c]{@{}c@{}}Perturb.\\ Ratio\end{tabular}} & \textbf{NCI1} & \textbf{D\&D} & \textbf{PROTEINS} \\ \hline
\multirow{5}{*}{\textbf{DIFFPOOL}}     & 0.0    &  62.32 ± 1.90 & 66.95 ± 2.41 & 68.20 ± 2.02\\ 
                                       & 0.5    &  49.48 ± 0.01  & 46.95 ± 0.06  & 50.84  ±  0.03 \\ 
                                       & 1.0    &   42.72 ± 0.02 & 46.27 ± 0.02 & 40.51  ±  0.02\\ 
                                       & 1.5    &  39.70 ± 0.02  & 45.93 ± 0.02 &  37.96  ±  0.023\\ 
                                       & 2.0    &  35.09 ± 0.03  & 45.83 ± 0.06 &  36.44  ±  0.05 \\ \hline
\multirow{5}{*}{\textbf{ASAP}}         & 0.0    & 71.48 ± 0.42 & 76.87 ± 0.7 & 74.19 ± 0.79 \\ 
                                       & 0.5    &  51.45±0.02  &66.78 ± 0.06  &  57.41 ± 0.04\\ 
                                       & 1.0    &  50.28±0.02  &64.11 ± 0.03 &  49.89 ± 0.03\\ 
                                       & 1.5    &   49.87±0.01 &63.82 ± 0.02 &  45.00 ± 0.04\\ 
                                       & 2.0    &  48.17±0.01  &61.28 ± 0.03   &  44.25 ± 0.04\\ \hline
\multirow{5}{*}{\textbf{\begin{tabular}[c]{@{}c@{}c@{}}HIBPool\end{tabular}}}                   & 0.0    & 83.45  ± 0.73  & 83.05 ± 0.34  &  82.88 ± 0.23\\ 
                                       & 0.5    &  81.50 ± 0.24 & 82.48 ± 1.59 & 81.38 ± 0.42  \\ 
                                       & 1.0    &  81.17 ± 0.09  & 81.35 ± 0.00 & 81.08 ± 0.00 \\ 
                                       & 1.5    &  79.73 ± 2.25 & 80.92 ± 0.42 &  81.08 ± 0.00 \\ 
                                       & 2.0    &  79.35 ± 1.90  &  80.5 ± 0.68 &  81.08 ± 0.00 \\ \hline
\end{tabular}
}
\caption{Feature Perturbation Analysis}
\label{tab:perturb_analysis}
\end{table}

\subsection{Robustness Against Feature Perturbations}
To observe the resilience of the model in feature attack, we perturb the input feature. We add gaussian noises to each dimension of node features as follows,
\begin{equation}
    \widehat{X^{(l)}} = X^{(l)} + \gamma \cdot r \cdot \epsilon
\end{equation}
where r is the mean of maximal value of each node's features and $\epsilon\sim\mathcal{N}(0,1)$ and $\gamma$ is noise-ratio. We test models' performance with $\gamma \in \{ 0.5, 1, 1.5, 2.0\}$. As reported in Table \ref{tab:perturb_analysis}, the performance of DIFFPOOL and ASAP degrade significantly when the input feature is perturbed compared to our model. These state-of-the-art models can not distinguish the noisy information from the input features, resulting in inferior performance when the input is perturbed. 

\subsection{Analysis on Multiresolution HIBPool (MHIBPool)}
We varied the value of $\alpha$ and observed results in Table~\ref{tab:baseline_comparion}. MHIBPool shows the best performance in the ENZYME, D\&D, NC1, and NC109 datasets, and in the rest of the datasets, its performance is competitive to HIBPool. 

 To evaluate how much the stability of communities discovered by the Louvain algorithm affects our model's performance, we detect communities once for each run. Then we take the average of the accuracy of all runs. The smaller standard deviations of accuracies across multiple runs indicate that our model consistently performs with the discovered communities for all graph datasets. 

\section{Conclusion}
We introduce HIBPool, a novel community structure-aware pooling method for graph classification. We leverage the Information Bottleneck principle into a novel DiP-Readout function to learn graph-level representation hierarchically that is minimal sufficient information for the task but maximal informative to target. IB helps HIBPool to attain resilience to feature perturbations. Empirical results exhibit the efficacy of HIBPool on all benchmark datasets. Further extensive analysis validates our findings, i.e., i) local pooling that is capable of important community structures are superior to global pooling, ii) a sophisticated pooling function is required to preserve distinguishable structures even it is indispensable for graphs with homogeneous node features, and iii) IB principle promotes pooling function to learn distinct representations. In future works, there are scopes to work on the limitations of community detection algorithms such as resolution limit, stability, etc.   
\section{Acknowledgements}
This project is supported by a grant from the Independent University Bangladesh and ICT Division of Bangladesh Government.

\bibliographystyle{IEEEtran}
\bibliography{ijcnn21}

\end{document}